\documentclass[runningheads]{llncs}

\usepackage[T1]{fontenc}
\usepackage{graphicx}
\usepackage{amsmath,amssymb}
\usepackage{booktabs}
\usepackage{microtype}
\usepackage[hidelinks]{hyperref}

\makeatletter
\renewcommand\paragraph{\@startsection{paragraph}{4}{\z@}%
                       {-12\p@ \@plus -4\p@ \@minus -4\p@}%
                       {-0.5em \@plus -0.22em \@minus -0.1em}%
                       {\normalfont\normalsize\bfseries\boldmath}}
\makeatother

\begin{document}

\title{Learning from Complementary Ultrasound Representations\\
for Liver Disease Classification}
\titlerunning{Complementary Ultrasound Representations for Liver Disease Classification}

\author{Sabahattin Mert Daloglu\inst{1} \and
Gokce Bekar\inst{1} \and
Ceren Coskun\inst{1} \and
Senanur Sahin\inst{1} \and
Harvey Castro\inst{1} \and
Soner Hacihaliloglu\inst{1} \and
Halley P. Letter\inst{2} \and
Ilker Hacihaliloglu\inst{1,3}}
\authorrunning{Daloglu et al.}
\institute{PONS Incorporated, Newark, NJ, USA\\
\email{md@ponstech.co, gb@ponstech.co, cc@ponstech.co, ss@ponstech.co,}\\
\email{harvery@ponstech.co, soner@ponstech.co}
\and
Department of Radiology, Mayo Clinic, Jacksonville, FL, USA\\
\email{letter.haley@mayo.edu}
\and
Department of Radiology, Department of Medicine, University of British Columbia, Vancouver, Canada\\
\email{ilker.hacihaliloglu@ubc.ca}}

\maketitle

\begin{abstract}
Differentiating non-alcoholic steatohepatitis (NASH) from non-alcoholic fatty liver disease (NAFLD) using ultrasound remains challenging due to subtle tissue alterations and the limited information available in conventional B-mode imaging. In this work, we investigate whether complementary ultrasound representations derived from the same acquisition can improve NASH versus NAFLD classification. Specifically, we combine conventional B-mode ultrasound with physics-guided and local phase-based image representations and evaluate their effectiveness using self-supervised masked autoencoders (MAEs) and graph convolutional networks (GCNs). Experiments were conducted on a multi-site Mayo Clinic cohort consisting of 2,547 liver ultrasound scans from 125 patients. Compared with conventional B-mode ultrasound alone, complementary ultrasound representations consistently improved classification performance, yielding gains of up to 32.4\% in accuracy (0.635$\rightarrow$0.841) and 91.2\% in F1-score (0.444$\rightarrow$0.849). Furthermore, performance improvements were consistently observed across age groups, sex, race, ethnicity, and acquisition sites.
\keywords{Liver Disease \and NASH \and NAFLD \and Ultrasound \and Machine Learning}
\end{abstract}

\section{Introduction}

Metabolic dysfunction-associated steatotic liver disease (MASLD), previously known as non-alcoholic fatty liver disease (NAFLD), is the most common chronic liver disease worldwide, affecting nearly one-third of the global population~\cite{ref1,ref2}. Its progressive subtype, metabolic dysfunction-associated steatohepatitis (MASH), formerly referred to as non-alcoholic steatohepatitis (NASH), can lead to fibrosis, cirrhosis, hepatocellular carcinoma, and liver failure if left untreated~\cite{ref3}. Early detection is therefore essential for enabling timely therapeutic intervention and preventing irreversible liver damage~\cite{ref4}. For consistency with the clinical labels used in this study and prior literature, we subsequently refer to these conditions as NAFLD and NASH.

Although liver biopsy remains the clinical reference standard for diagnosing and staging NASH, its invasive nature, sampling variability, procedural risks, and cost limit its use for large-scale screening and longitudinal monitoring~\cite{ref5}. Consequently, ultrasound has emerged as one of the most widely used imaging modalities for liver disease assessment due to its low cost, safety, portability, and widespread availability~\cite{ref4,ref5}. However, accurate ultrasound-based classification of NAFLD and NASH remains challenging. Conventional B-mode ultrasound images are affected by speckle noise, low contrast, acoustic artifacts, operator dependence, and variability across imaging systems and acquisition protocols~\cite{ref4,ref5,ref6,ref7,ref8,ref9}. Moreover, early-stage disease is often characterized by subtle parenchymal changes that may not be readily visible on standard B-mode imaging, limiting diagnostic sensitivity and contributing to variability in clinical interpretation~\cite{ref4}.

To address the limitations of conventional ultrasound imaging, numerous artificial intelligence (AI) approaches have been proposed for automated liver disease assessment. Deep learning methods, including convolutional neural networks (CNNs), transformers, radiomics-based frameworks, and self-supervised learning models, have demonstrated promising performance for steatosis detection, fibrosis staging, and liver disease classification~\cite{ref6,ref7,ref8,ref9,ref10,ref11,ref12}. Prior studies have explored CNN-based approaches for fatty liver disease detection and NAFLD classification from B-mode ultrasound images~\cite{ref6,ref7,ref8,ref9,ref10,ref11,ref12}, while others have incorporated enhanced ultrasound representations to improve classification performance using CNNs~\cite{ref9}. Despite these advances, differentiating NASH from NAFLD remains particularly challenging because disease-related imaging differences are often subtle and inadequately captured by conventional B-mode ultrasound. Moreover, existing studies~\cite{ref6,ref7,ref8,ref9,ref10,ref11,ref12} have predominantly employed supervised CNN-based architectures focused on steatosis grading or broad liver disease classification, with no prior work investigating ultrasound-based NASH versus NAFLD differentiation as a dedicated diagnostic task.

\begin{figure}[t]
\centering
\includegraphics[width=0.72\textwidth]{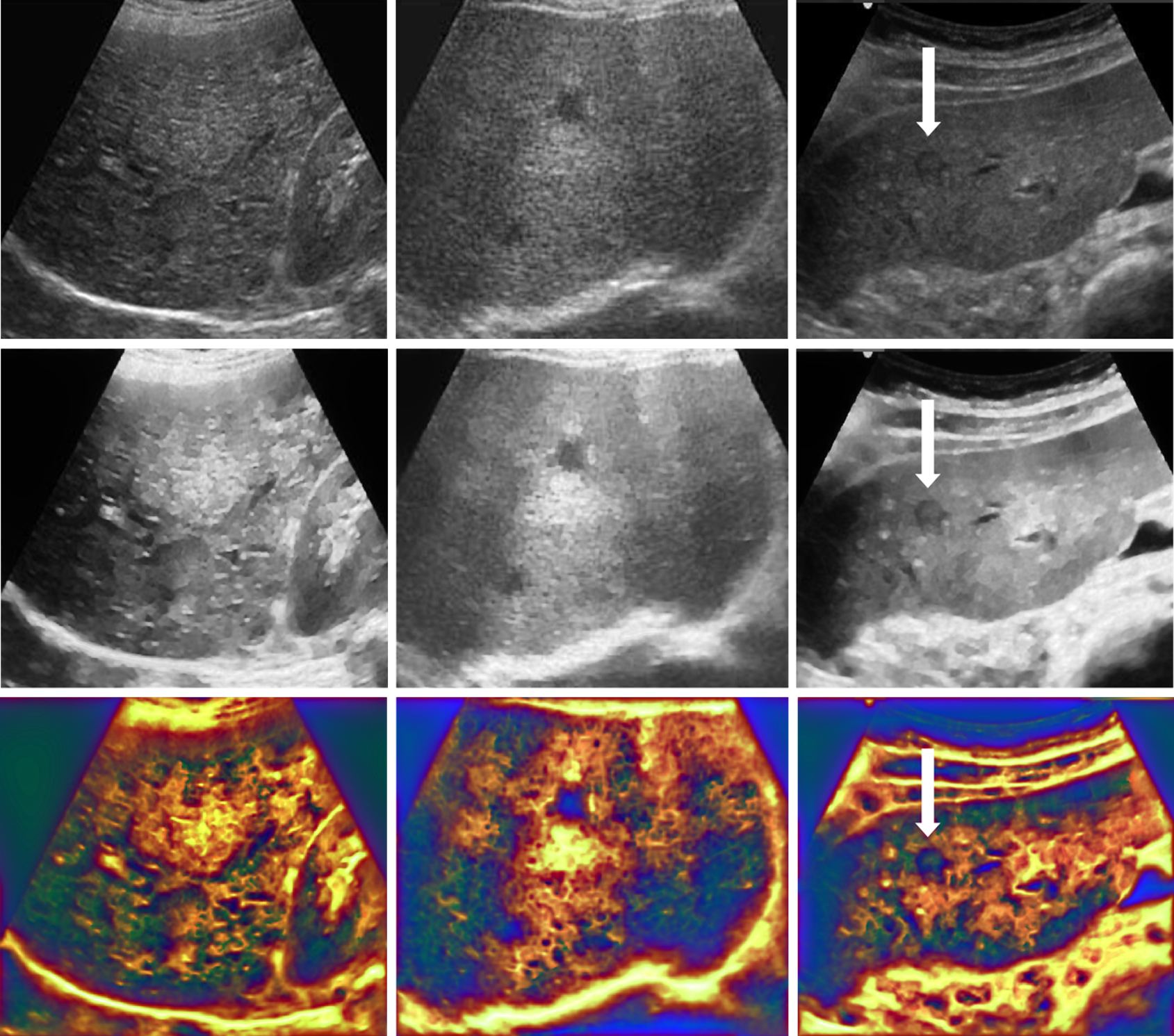}
\caption{Original liver ultrasound data (BUS) top row, and the enhanced representations QiUS (middle row), and EnUS (last row).}
\label{fig:cur}
\end{figure}

In this work, we investigate the use of complementary ultrasound representations for differentiating NASH from NAFLD. Specifically, we integrate conventional B-mode imaging with physics-guided and local phase-based image representations and evaluate their effectiveness using two learning paradigms that have not previously been explored for this task: self-supervised masked autoencoders (MAEs)~\cite{ref13} and graph convolutional networks (GCNs)~\cite{ref14}. The proposed framework is evaluated on a multi-site Mayo Clinic cohort comprising 2,547 liver ultrasound scans from 125 patients. Experimental results demonstrate consistent performance improvements over conventional B-mode ultrasound alone across demographic subgroups and acquisition sites. Our contributions include: 1-~We present the first study investigating complementary ultrasound representations for AI-based differentiation of NASH and NAFLD. 2-~We evaluate, for the first time, MAEs and GCNs for ultrasound-based NASH versus NAFLD classification. 3-~We perform a comprehensive multi-site evaluation, including subgroup analyses across age, sex, race, ethnicity, and acquisition site, demonstrating consistent improvements over conventional B-mode ultrasound imaging.

\section{Patients and Methods}

\subsection{Data Sources and Study Population}
De-identified DICOM ultrasound examinations were organized at the patient level and linked to corresponding clinical and pathological records. The reference standard diagnosis was established using liver biopsy results. The dataset, provided by the Mayo Clinic, consists of 2,547 liver ultrasound scans acquired from 125 patients, including 1,274 NAFLD scans and 1,273 NASH scans collected from multiple Mayo Clinic sites. Some patients were referred to Mayo Clinic after undergoing ultrasound examinations at external healthcare institutions.

\subsection{Complementary Ultrasound Representations (CUR)}

To provide complementary information for downstream learning, each ultrasound examination was represented using three image modalities: conventional B-mode ultrasound (BUS), a quality-improved ultrasound representation (QiUS), and an enhanced ultrasound representation (EnUS) (Fig.~\ref{fig:cur}).

\paragraph{Quality-Improved Ultrasound (QiUS).}
QiUS images are generated using a physics-guided attenuation characterization framework that models ultrasound signal propagation within tissue. The method estimates spatially varying transmission and attenuation characteristics and applies contextual regularization to compensate for signal degradation caused by acoustic attenuation. The resulting representation enhances attenuation-dependent tissue information while reducing acquisition-related artifacts and image inconsistencies commonly encountered in ultrasound imaging (Fig.~\ref{fig:cur}). The concept was previously utilized for breast, bone shadow and needle shaft/tip enhancement from ultrasound data~\cite{ref16,ref17,ref18} and is based on the patented work of~\cite{ref15}. The regularization parameter was set to 2, consistent with previous studies~\cite{ref15,ref16,ref17}. The tissue echogenicity constant was defined as 80\% of the maximum intensity value of the corresponding B-mode ultrasound (BUS) image, following the methodology described in~\cite{ref15,ref16,ref17}.

\paragraph{Enhanced Ultrasound (EnUS).}
EnUS images are generated through local phase-based feature extraction in the frequency domain through band-pass quadrature filtering. Specifically, alpha-scale space derivative (ASSD) filtering is employed~\cite{ref9,ref15,ref16,ref17}, to BUS image, to derive local phase representations that emphasize structural and morphological information while remaining relatively invariant to image intensity, scanner settings, and acoustic gain~\cite{ref9,ref15,ref16,ref17}. Consequently, the EnUS representation enhances tissue interfaces, parenchymal texture, and anatomical boundaries that may not be readily visible in conventional B-mode images (Fig.~\ref{fig:cur}). The ASSD filter is defined as
\begin{equation}
\mathrm{ASSDF}(\omega)=
\begin{cases}
n_{c}\,\omega^{a}\exp\!\bigl(-(\sigma\omega)^{2\alpha}\bigr) & \omega>0,\\[2pt]
0 & \text{otherwise.}
\end{cases}
\label{eq:assd}
\end{equation}
\noindent where $n_{c}$ is a unit normalization constant calculated from the filter $\alpha$ value. During this work following parameters were used: filter alpha-scale parameter $\sigma = 25$~pixels, $a = 1.83$ and constant derivative parameter $\alpha = 0.2$. During $n_{c}$ calculation we have used number of scales $s = 2$ and number of orientations $= 6$ ($30^{\circ}$ increments starting from $0^{\circ}$).

Complete methodological details and validation of the QiUS and EnUS generation processes can be found in previous publications~\cite{ref9,ref15,ref16,ref17}.

The QiUS and EnUS representations provide complementary views of the same ultrasound acquisition. Whereas QiUS emphasizes acoustic propagation and attenuation-related characteristics, EnUS captures structural and morphological information derived from local phase analysis (Fig.~\ref{fig:cur}). Together with the original BUS image, these representations form a set of complementary ultrasound representations (CUR) that provide distinct yet synergistic descriptions of tissue appearance, enabling downstream learning models to leverage both acoustic and structural information for liver tissue characterization.

\begin{figure}[t]
\centering
\includegraphics[width=\textwidth]{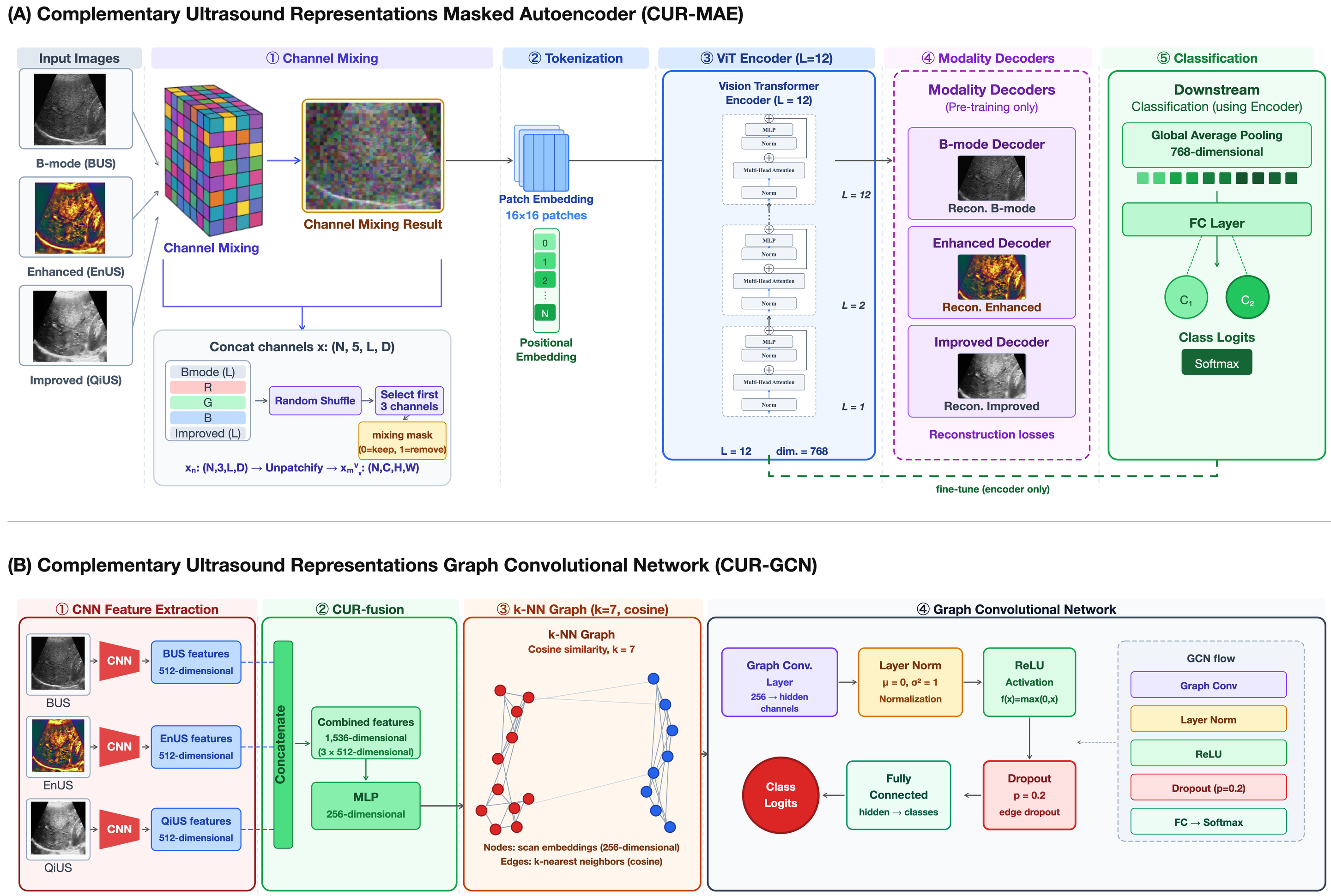}
\caption{Overview of the proposed learning frameworks for NASH versus NAFLD classification using CUR. (A)~CUR-MAE employs self-supervised masked autoencoding to learn joint representations from BUS, EnUS, and QiUS images. (B)~CUR-GCN extracts and fuses features from BUS, EnUS, and QiUS images and utilizes graph convolutional learning to model relationships among examinations for classification.}
\label{fig:framework}
\end{figure}

\subsection{Learning with Complementary Ultrasound Representations}

To evaluate the proposed CUR approach, we investigate two learning paradigms. First, a self-supervised masked autoencoder (CUR-MAE) is employed to learn robust representations from BUS, QiUS, and EnUS inputs. Second, a graph convolutional network (CUR-GCN) is utilized to model relationships among ultrasound examinations through graph-based learning. Both models leverage the complementary acoustic and structural information encoded within the CUR representations for NASH versus NAFLD classification.

\subsubsection{CUR-Masked Autoencoder (CUR-MAE) Network} 

As illustrated in Figure~\ref{fig:framework}, the three ultrasound representations derived from the same acquisition are first combined through a CUR Mixing module inspired by channel-mixing operations~\cite{ref19}. The channels from BUS, EnUS, and QiUS images are concatenated, randomly shuffled within each image patch, and partially sampled to generate mixed CUR-feature inputs (Fig.~\ref{fig:framework}). This process encourages the network to learn feature interactions across representations while reducing dependence on any single image type. The resulting mixed images are partitioned into $16\times16$ patches and embedded into a latent feature space using a Vision Transformer (ViT) encoder. Positional embeddings are added to preserve spatial information, and the encoder processes the patch sequence through 12 transformer blocks comprising multi-head self-attention, multilayer perceptrons, normalization layers, and residual connections (Fig.~\ref{fig:framework}). The self-attention mechanism enables the model to capture long-range dependencies and relationships among anatomical structures across the entire image. During self-supervised pre-training, randomly masked image regions are reconstructed, forcing the encoder to learn informative structural, textural, and acoustic characteristics shared across the BUS, EnUS, and QiUS representations. After pre-training, the decoder is discarded and only the learned encoder is retained for downstream diagnostic classification. Final classification is performed using global average pooling followed by a fully connected classification layer.

To improve representation learning and mitigate the limited availability of labeled liver ultrasound data, the MAE was first pretrained on more than 230,000 publicly available de-identified ultrasound images and videos collected from multiple open-source datasets~\cite{ref20,ref21,ref22,ref23,ref24,ref25,ref26,ref27,ref28,ref29}. For CUR-MAE, each ultrasound acquisition was represented using three complementary image modalities (BUS, QiUS, and EnUS), effectively tripling the number of training representations available during pretraining and enabling the model to learn richer acoustic and structural features.
\subsubsection{CUR-Graph Convolutional Network (CUR-GCN) Architecture}
Unlike conventional image-based classifiers that process each examination independently, CUR-GCN explicitly models relationships among examinations by representing them as nodes in a similarity graph.

For each ultrasound representation, a dedicated CNN encoder extracts a 512-dimensional feature vector. The resulting BUS, EnUS, and QiUS features are concatenated to form a 1,536-dimensional multi-feature representation, which is subsequently compressed through a multilayer perceptron (MLP) into a unified 256-dimensional embedding. This fused representation serves as the node descriptor for graph construction (Fig.~\ref{fig:framework}).

A $k$-nearest-neighbor ($k$-NN) graph is then generated using cosine similarity between fused feature vectors, with $k = 7$. This value was selected through preliminary validation experiments to balance local neighborhood consistency and global graph connectivity, enabling effective information propagation across clinically similar examinations. By incorporating information from all three ultrasound representations during graph construction, the resulting graph exhibits improved homophily compared with graphs generated from a single image representation, thereby reducing fragmentation caused by image variability and noise.

CUR-GCN classification network consists of a graph convolution layer followed by layer normalization, ReLU activation, dropout, and a fully connected classification layer. During training, edge dropout is applied to improve robustness and reduce overfitting~\cite{ref14}. Through graph-based message passing, CUR-GCN aggregates information from neighboring examinations, enabling more reliable classification than single-feature graph learning approaches.

\section{Results}

\subsection{Implementation Details}
Model performance was evaluated using patient-level three-fold cross-validation. To prevent data leakage, all ultrasound examinations from a given patient were assigned exclusively to a single fold. Stratified sampling was employed to preserve the NASH and NAFLD class distributions across the training, validation, and test subsets. Within each fold, patients were split into training (70\%), validation (10\%), and testing (20\%) sets, and identical data partitions were used for all models to ensure fair comparison. Performance metrics were reported as the mean and standard deviation across three-fold cross-validation folds. To assess robustness across patient populations and acquisition environments, subgroup analyses were performed based on age, sex, race, ethnicity, and acquisition site; accuracy for each subgroup was computed by pooling patient-level predictions from the held-out test sets across all folds. The CUR-GCN model was trained for 125 epochs using cross-entropy loss and the AdamW optimizer, with the model achieving the highest validation accuracy selected for final evaluation. The CUR-MAE model was initialized using pretrained weights and fine-tuned for 10 epochs. A limited number of fine-tuning epochs was selected to mitigate overfitting, as longer training resulted in increased validation loss. The initial learning rate was set to $2\times10^{-4}$ and reduced by a factor of 10 using a step-based schedule. Additional regularization included weight decay (0.05) and layer-wise learning rate decay (0.75). Model parameters were updated after each iteration without gradient accumulation. All experiments were implemented in Python using PyTorch (v2.7.0). Training was performed on NVIDIA A100 GPUs (80~GB) and TPUs (128~GB) with CUDA~12.6 support. Automatic Mixed Precision (AMP) was employed to improve computational efficiency and reduce memory consumption. As a baseline comparison, we additionally implemented a CNN-EBV architecture as a baseline~\cite{ref9,ref30}, which utilizes the same complementary ultrasound representations (CUR) and an Equiangular Basis Vector (EBV) classifier~\cite{ref30}.

\subsection{Quantitative Results}
Table~\ref{tab:results} summarizes the classification performance of the investigated models. Across all architectures, incorporating CUR consistently improved classification performance compared with conventional B-mode ultrasound. The largest gains were observed for CUR-GCN, which improved accuracy, F1-score, and AUC by 32.4\%, 91.2\%, and 11.8\%, respectively. CUR-MAE also demonstrated substantial improvements, achieving gains of 27.1\% in accuracy and 33.8\% in F1-score. Analysis of sensitivity and specificity demonstrated that CUR improved disease discrimination across architectures. CUR-GCN achieved the largest sensitivity gain (+111.9\%), while CUR-MAE improved both sensitivity (+36.0\%) and specificity (+14.7\%), suggesting improved identification of both NASH and NAFLD cases. These findings indicate that complementary ultrasound representations provide discriminative information beyond conventional B-mode imaging and improve the differentiation of NASH and NAFLD. Unlike the MAE and GCN frameworks, the CNN-EBV baseline primarily exhibited a tradeoff between sensitivity and specificity following incorporation of CUR, resulting in unchanged F1-score despite improved accuracy. This observation suggests that the benefits of complementary ultrasound representations depend on the downstream learning architecture, with representation-learning and graph-based models appearing better suited to exploit the additional acoustic and structural information encoded within CUR.

\begin{table}[t]
\centering
\caption{Classification performance of BM and CUR-based learning frameworks. Results are averaged over all folds.}
\label{tab:results}
\resizebox{\textwidth}{!}{%
\begin{tabular}{@{}lccccc@{}}
\toprule
Model & Accuracy & AUC & F1 & Specificity & Sensitivity \\
\midrule
BM-GCN~\cite{ref14} & $0.635\pm0.102$ & $0.706\pm0.269$ & $0.444\pm0.385$ & $\mathbf{0.750\pm0.250}$ & $0.472\pm0.411$ \\
CUR-GCN & $\mathbf{0.841\pm0.045}$ & $\mathbf{0.789\pm0.217}$ & $\mathbf{0.849\pm0.045}$ & $0.683\pm0.161$ & $\mathbf{1.000\pm0.000}$ \\
\midrule
BM-MAE~\cite{ref19} & $0.643\pm0.211$ & $0.715\pm0.344$ & $0.622\pm0.204$ & $0.617\pm0.126$ & $0.694\pm0.337$ \\
CUR-MAE & $\mathbf{0.817\pm0.161}$ & $\mathbf{0.733\pm0.295}$ & $\mathbf{0.832\pm0.147}$ & $\mathbf{0.708\pm0.260}$ & $\mathbf{0.944\pm0.096}$ \\
\midrule
BM-CNN-EBV~\cite{ref9,ref30} & $0.454\pm0.042$ & $0.665\pm0.160$ & $\mathbf{0.541\pm0.077}$ & $0.250\pm0.250$ & $\mathbf{0.750\pm0.250}$ \\
CUR-CNN-EBV & $\mathbf{0.624\pm0.265}$ & $\mathbf{0.674\pm0.246}$ & $0.541\pm0.303$ & $\mathbf{0.725\pm0.198}$ & $0.528\pm0.411$ \\
\bottomrule
\end{tabular}%
}
\end{table}

\begin{figure}[t]
\centering
\includegraphics[width=\textwidth]{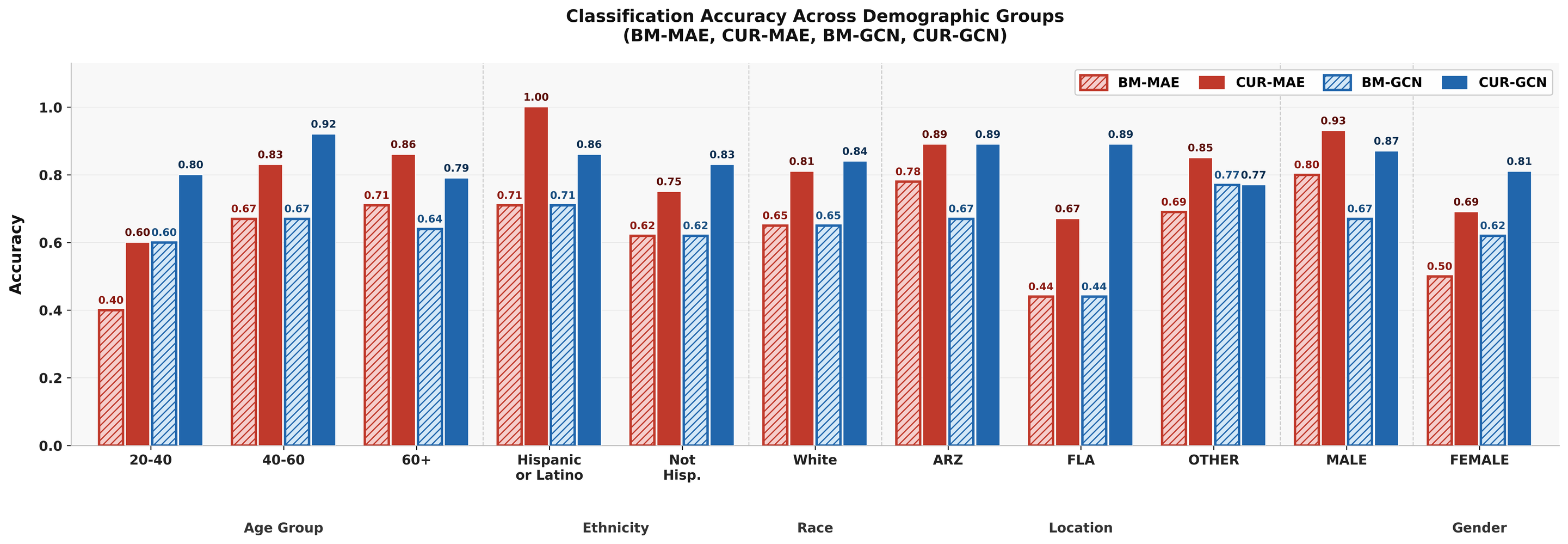}
\caption{Classification accuracy across demographic subgroups and acquisition sites for BUS and CUR using MAE and GCN models. CUR consistently improves performance over BM across most age, ethnicity, race, location, and sex subgroups.}
\label{fig:metadata}
\end{figure}

\subsection{Metadata Analysis}
Figure~\ref{fig:metadata} summarizes classification accuracy across demographic subgroups and acquisition sites. Overall, incorporating complementary ultrasound representations (CUR) consistently improved performance relative to conventional B-mode ultrasound across nearly all age, ethnicity, race, location, and sex categories. The largest gains were observed for the GCN-based framework, particularly among patients aged 40--60 years (0.67$\rightarrow$0.92), patients scanned at the Florida site (0.44$\rightarrow$0.89), and female patients (0.62$\rightarrow$0.81). Similarly, CUR-MAE demonstrated consistent improvements across all evaluated subgroups, achieving notable gains in Hispanic or Latino patients (0.71$\rightarrow$1.00), male patients (0.80$\rightarrow$0.93), and examinations acquired at the Florida site (0.44$\rightarrow$0.67). Importantly, no demographic subgroup exhibited a substantial performance degradation following incorporation of CUR, suggesting that the complementary representations provide robust and generalizable information across diverse patient populations and acquisition environments. These findings indicate that the benefits of CUR extend beyond overall classification performance and remain consistent across demographic and site-specific strata.

\section{Discussion and Future Work}

This study investigated the use of complementary ultrasound representations (CUR) for differentiating NASH from NAFLD using self-supervised MAEs and GCNs. Across all evaluated architectures, CUR consistently improved classification performance relative to conventional B-mode ultrasound, with the largest gains observed for CUR-GCN. Specifically, CUR-GCN achieved improvements of 32.4\% in accuracy, 91.2\% in F1-score, and 11.8\% in AUC, while CUR-MAE improved accuracy and F1-score by 27.1\% and 33.8\%, respectively. Analysis of operating characteristics further revealed that CUR-GCN achieved the largest sensitivity improvement (+111.9\%), whereas CUR-MAE improved both sensitivity (+36.0\%) and specificity (+14.7\%), indicating more balanced discrimination between NASH and NAFLD. Furthermore, metadata analysis demonstrated consistent performance improvements across age, sex, race, ethnicity, and acquisition-site subgroups, suggesting that the benefits of CUR are not restricted to a particular patient population or imaging environment.

Several limitations should be acknowledged. First, although the dataset comprised 2,547 ultrasound scans, these data were acquired from only 125 patients. Consequently, some demographic subgroups contained relatively small numbers of patients, and subgroup-specific results should therefore be interpreted with caution. While certain subgroups achieved near-perfect performance, these observations require validation in larger and more balanced cohorts. Second, several evaluation metrics exhibited relatively large standard deviations across cross-validation folds, particularly for AUC. This variability likely reflects the limited cohort size and the inherent heterogeneity of liver ultrasound imaging. Although consistent improvements were observed across architectures and demographic groups, larger cohorts are needed to obtain more stable estimates of model performance. Third, the study was conducted retrospectively using data from a single healthcare system. Although the cohort included examinations acquired across multiple Mayo Clinic locations and a subset of patients referred from external institutions, additional external validation is required to assess generalizability across different scanners, imaging protocols, and patient populations.

Future work will focus on expanding the study through larger multi-institutional cohorts and external validation datasets as well as investigating multimodal learning frameworks that combine complementary ultrasound representations with language-based features extracted from radiology reports to further improve non-invasive NASH versus NAFLD classification. In addition, future studies will explore the utility of the proposed framework for fibrosis staging, longitudinal disease monitoring, and risk stratification. These efforts will help establish the clinical utility of complementary ultrasound representations for non-invasive liver disease assessment and support the development of robust AI tools for differentiating NASH from NAFLD.

\paragraph{Acknowledgement.}
Research reported in this work was supported by the National Institute on Minority Health and Health Disparities of the National Institutes of Health under award number 1R43MD020006-01A1. The content is solely the responsibility of the authors and does not necessarily represent the official views of the National Institutes of Health.
We would like to thank the Mayo Clinic and the Mayo Accelerate Program for their valuable support and collaboration throughout this project. Their guidance, expertise, and provision of data were instrumental in the development and evaluation of our complementary ultrasound representation learning framework.

\paragraph{Ethics Statement.}
This study involves analysis of de-identified
data via the Mayo Clinic Discover Platform.
In accordance with the Code of Federal Regulations, 45 CFR 46.102, the noted activity
does not require IRB review. Data shown
and reported in this manuscript have been
extracted from this environment using an
established protocol for data extraction,
aimed at preserving patient privacy. The
data has been de-identified pursuant to an
expert determination in accordance with the
HIPAA Privacy Rule. Any data beyond what
is reported in the manuscript, including but
not limited to the raw electronic health record
data, cannot be shared or released due to the parameters of the expert determination to
maintain the data de-identification.

\end{document}